# Boosting LiDAR-based Semantic Labeling by Cross-Modal Training Data Generation

Florian Piewak[*,1,2], Peter Pinggera[1],
Manuel Schäfer[1], David Peter[1], Beate Schwarz[1], Nick Schneider[1],
David Pfeiffer[**,1], Markus Enzweiler[1], and Marius Zöllner[2,3]

[1]Daimler AG, R&D, Stuttgart, Germany
[2]Karlsruhe Institute of Technology (KIT), Karlsruhe, Germany
[3]Forschungszentrum Informatik (FZI), Karlsruhe, Germany

**Abstract.** Mobile robots and autonomous vehicles rely on multi-modal sensor setups to perceive and understand their surroundings. Aside from cameras, LiDAR sensors represent a central component of state-of-the-art perception systems. In addition to accurate spatial perception, a comprehensive semantic understanding of the environment is essential for efficient and safe operation. In this paper we present a novel deep neural network architecture called *LiLaNet* for point-wise, multi-class semantic labeling of semi-dense LiDAR data. The network utilizes virtual image projections of the 3D point clouds for efficient inference. Further, we propose an automated process for large-scale cross-modal training data generation called *Autolabeling*, in order to boost semantic labeling performance while keeping the manual annotation effort low. The effectiveness of the proposed network architecture as well as the automated data generation process is demonstrated on a manually annotated ground truth dataset. *LiLaNet* is shown to significantly outperform current state-of-the-art CNN architectures for LiDAR data. Applying our automatically generated large-scale training data yields a boost of up to 14 percentage points compared to networks trained on manually annotated data only.

## 1 Introduction

Within the fields of mobile robotics and autonomous driving, vehicles are typically equipped with multiple sensors of complementary modalities such as cameras, LiDAR and RADAR in order to generate a comprehensive and robust representation of the environment [2–5]. Each sensor modality leverages its specific strengths to extract as much information as possible from the observed scene. Based on the combined sensor data, a detailed environment model, for example in the form of a dynamic occupancy grid map [6], is created. This environment model provides the basis for high-level tasks such as object tracking [7], situation analysis and path planning [8]. In order to master these high-level tasks, it is essential for an autonomous vehicle to not only distinguish between generic obstacles and free-space, but to also obtain a deeper semantic understanding of its surroundings. Within the field of computer vision, the corresponding task of semantic image labeling has experienced a significant boost in recent years due to the resurgence of advanced deep learning techniques [9]. However, detailed semantic information of similar quality has to be extracted independently from each of the sensor modalities to maximize system performance, availability and safety. Therefore, in this paper we introduce *LiLaNet*, an efficient deep neural network architecture for point-wise, multi-class semantic labeling of semi-dense LiDAR data.

---

[*] Corresponding author (florian.piewak@daimler.com)
[**] Contributed while with Daimler AG

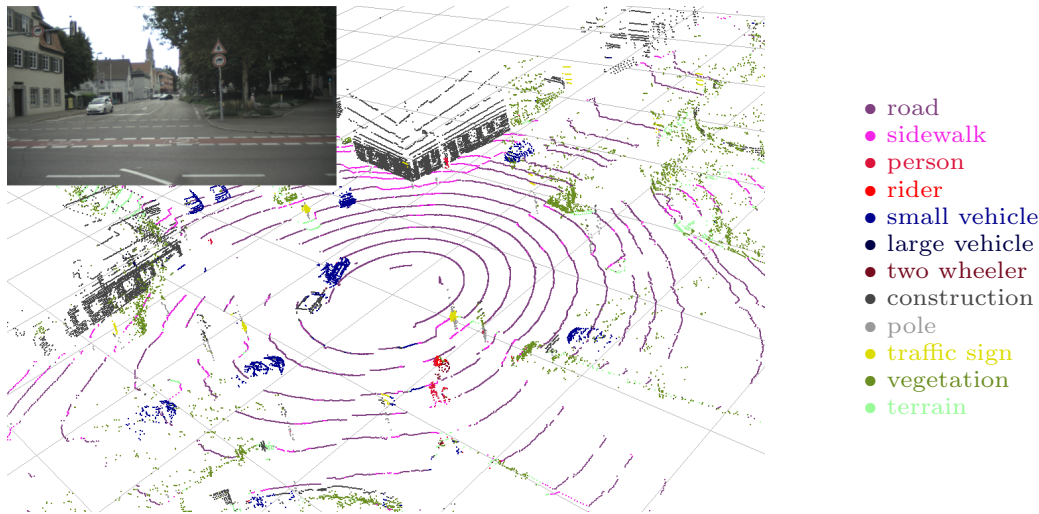

**Fig. 1.** Example of a LiDAR point cloud labeled by *LiLaNet*. Point colors correspond to the Cityscapes semantic class color coding [1]. The test vehicle is headed to the top right, the corresponding camera image is shown on the top left for clarity.

As has been well established within the camera domain, large-scale datasets are of paramount importance for training competitive deep neural networks. Consequently, large-scale generic datasets such as ImageNet [10] and COCO [11], as well as medium-scale datasets dedicated to road scenarios such as KITTI [12] and Cityscapes [1] have been made available for this purpose. In contrast, within the LiDAR domain only indoor datasets [13–15] or outdoor datasets [16] obtained from high-resolution stationary sensors have been published to date. For this reason some authors [17–19] have resorted to the indirect extraction of the desired LiDAR point semantics from the KITTI dataset, using annotated object bounding boxes or the camera road detection benchmark. While this type of indirect extraction eventually yields viable training datasets, it is relatively cumbersome and limited to only a small set of semantic classes. Hence, here we propose a so-called *Autolabeling* process, an effective approach for the automated generation of large amounts of semantically annotated mobile LiDAR data by the direct transfer of high-quality semantic information from a registered reference camera image. The semantic information in the reference image is obtained using an off-the-shelf neural network. We show that the datasets obtained with this approach significantly boost the LiDAR-based semantic labeling performance, in particular when augmented with a small manually annotated dataset for fine-tuning.

Our main contributions can be summarized as follows:

1. An efficient Convolutional Neural Network (CNN) architecture for high-quality semantic labeling of semi-dense point clouds, as provided by state-of-the-art mobile LiDAR sensors.
2. A large-scale automated cross-modal training data generation process for boosting the LiDAR-based semantic labeling performance.
3. A thorough quantitative evaluation of the semantic labeling performance, including an analysis of the proposed automated training data generation process.

## 2   Related Work

LiDAR-based semantic labeling has gained increased attention in recent years due to the availability of improved mobile sensor technology, providing higher resolution and longer range at reduced cost. The various proposed approaches of LiDAR-based semantic labeling can be discriminated by the way the point-wise 3D information is utilized.

The first approaches using depth information for semantic labeling were based on RGB-D data, which complements RGB image data with an additional depth channel [20, 21], allowing to recycle 2D semantic image labeling algorithms. Often a stereo camera was used to create a dense depth image, which was then fused with the RGB image. Tosteberg [22] developed a technique to use depth information of a LiDAR sensor accumulated over time to project it into the camera space. The accumulation yields a depth image of increased density without requiring dedicated upsampling algorithms.

A different category of approaches considers the 3D LiDAR data as an unordered point cloud, including PointNet [23], PointNet++ [24] and PointCNN [25]. The PointNet architecture [23] combines local point features with globally extracted feature vectors, allowing for the inference of semantic information on a point-wise basis. Extending this idea, PointNet++ [24] introduces a hierarchical PointNet architecture to generate an additional mid-level feature representation for an improved handling of point neighborhood relations. Both approaches are evaluated successfully on indoor scenes but reach their limits in large scale outdoor scenarios. The PointCNN [25] approach is based on unordered point clouds as well, but introduces modified convolution layers extended by permutations and weighting of the input features. This allows to transfer the advantages of traditional CNNs to unordered point cloud processing. However, the approach is only used for object detection and has not yet been applied to semantic labeling of point clouds.

Yet another way of representing LiDAR input data is within cartesian 3D space, which is used in the SEGCloud [26] and OctNet [27] methods. Here a Voxel (SEGCloud) or an OctTree (OctNet) representation is created and the convolution layers are extended to 3D convolutions. These approaches retain the original 3D structure of the input points, making them more powerful in preserving spatial relations. However, the algorithms have to cope with the high sparsity of the data, and inference time as well as memory requirements increase drastically for large-scale outdoor scenes.

A possible solution to avoid the computational complexity of 3D convolutions is the rendering of 2D views of the input data. Based on such 2D views, state-of-the-art image-based deep learning algorithms can be applied. Depending on the use case, different viewpoints or virtual cameras may be used. Caltagirone et al. [19] use a top-view image of a LiDAR point cloud for labeling road points within a street environment. This top-view projection of the LiDAR points is a valid choice for road detection, but the resulting mutual point occlusions generate difficulties for more general semantic labeling task as in our case. An alternative is to place the virtual camera origin directly within the sensor itself. The resulting 2D view is often visualized via a cylindrical projection of the LiDAR points (see Fig. 2), which is particularly suitable for the regular measurement layout of common rotating LiDAR scanners. In this case, the sensor view provides a dense depth image, which is highly advantageous for subsequent processing steps. Wu et al. [17] uses this type of input image for the SqeezeSeg architecture, which performs a SqeezeNet-based [28] semantic labeling to segment cars, pedestrians and cyclist. The bounding boxes of the KITTI object detection dataset [12] are used to transfer the point-wise semantic information required for training and evaluation. The approach of Dewan et al. [18] uses the cylindrical projection of the LiDAR data as an input for a CNN based on the Fast-Net architecture [29] to distinguish between movable and non-movable points. Similar to Wu et al [17], the KITTI object detection dataset is used for transferring the ground-truth bounding box labels to the enclosed points.

Varga et al [30] propose an alternative method to generate a semantically labeled point cloud at runtime, based on a combined setup of fisheye cameras and LiDAR sensors. First, pixel-wise semantics are extracted from the camera images via a CNN model trained on Cityscapes [1]. Subsequently, the LiDAR points are projected into the images to transfer the semantic information from pixels to 3D points. However, no semantic information is inferred on the LiDAR point cloud itself, and spatial and temporal registration of the sensor modalities remains a challenge. In the present paper we take the idea of [30] one step further and utilize a joint camera/LiDAR sensor setup to generate large amounts of 3D semantic training data. The data is then used to train a deep neural network to infer point-wise semantic information directly on LiDAR data. We show that combining the large amounts of automatically generated data with a small manually annotated dataset boosts the overall semantic labeling performance significantly.

## 3   Method

We introduce a novel CNN architecture called *LiLaNet* for the point-wise, multi-class semantic labeling of LiDAR data. To obtain high output quality and retain efficiency at the same time, we aim to transfer lessons learned from image-based semantic labeling methods to the LiDAR domain. The cylindrical projection of a 360° point cloud captured with a state-of-the-art rotating LiDAR scanner is used as input to our networks. Training is boosted by an efficient automated cross-modal data generation process, which we refer to as *Autolabeling*.

### 3.1   LiDAR Images

LiDAR sensors measure the time of flight of emitted laser pulses in order to determine the distance of surrounding objects with high accuracy. In addition, modern sensors even provide coarse reflectivity estimates on a point-wise basis. For the following experiments we consider a Velodyne VLP32C LiDAR scanner, which features 32 vertically stacked send/receive modules rotating around a common vertical axis. While rotating, each module periodically measures the distance and reflectivity at its current orientation, i.e. at the respective azimuth and elevation angles. We combine the measurements of a full 360° scan to create cylindrical depth and reflectivity images, as illustrated in Fig. 2. This projection represents the view of a virtual 360° cylindrical camera placed at the sensor origin. At ten revolutions per second, images of size $1800 \times 32$ pixels are obtained.

The cylindrical point cloud projection provides dense depth and reflectivity images which are free from mutual point occlusions. This allows for the application of optimized 2D convolution layers, as used with great success in state-of-the-art image-based CNN architectures. In this way, inference time is reduced drastically compared to the use of full 3D input representations such as voxel grids or octtrees. Further, since measurement times and orientation angles are known with high accuracy, it is straightforward to transform the cylindrical image back into a full three-dimensional point cloud representation without any loss of information.

In cases where no laser reflection is received by the sensor, for example when pointed towards the sky, pixels corresponding to the respective measurement angles are marked as invalid in the resulting depth image.

### 3.2   Class Mapping

Based on initial experiments we conducted with regard to the discrimination of different semantic classes in LiDAR data, we first apply a mapping of the original Cityscapes labelset [1] to a reduced set of 13 semantic classes (see Table 1). This reduced label set is better tailored to the specific properties of the data provided by current LiDAR sensors, where limited spatial resolution and coarse reflectivity estimates prohibit truly fine-grained semantic differentiation. For example, the original 'truck' and 'bus' classes are merged into a common 'large vehicle' class. Similarly, the original 'motorcycle' and 'bicycle' classes are combined into a single

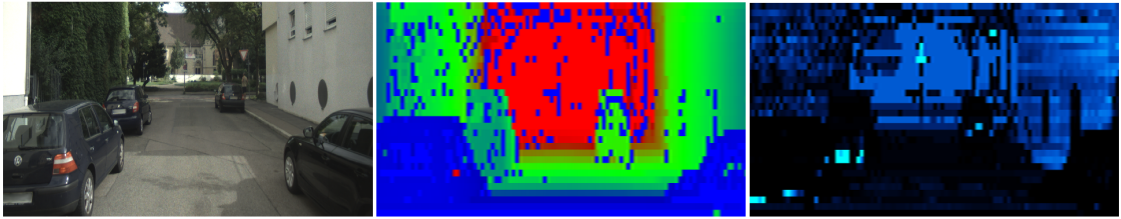

**Fig. 2.** Example of a depth image (center, blue = close, red = far) and reflectivity image (right, black = non-reflective, cyan = highly reflective) resulting from the cylindrical point cloud projection. The corresponding camera image is shown on the left. Note that here the 360° LiDAR scan has been cropped to the camera field of view for illustration.

| Cityscapes | LiDAR Semantics | | Cityscapes | LiDAR Semantics |
|---|---|---|---|---|
| road | road | | building | construction |
| sidewalk | sidewalk | | wall | construction |
| person | person | | fence | unlabeled |
| rider | rider | | pole | pole |
| car | small vehicle | | traffic sign | traffic sign |
| truck | large vehicle | | traffic light | construction |
| bus | large vehicle | | vegetation | vegetation |
| on rails | large vehicle | | terrain | terrain |
| motorcycle | two wheeler | | sky | sky |
| bicycle | two wheeler | | | |

**Table 1.** Mapping of the Cityscapes label set [1] to the reduced LiDAR label set

'two wheeler' class. The 'fence' class is not retained in our mapping, as such thin and porous structures are hardly captured by LiDAR sensors. Note that the reduced label set still provides an abundance of valuable semantic information with adequate detail, which is highly beneficial for a large set of application scenarios.

### 3.3 LiLaNet Network Architecture

Using the LiDAR images described in Section 3.1 as input, we present a dedicated CNN architecture for high-quality LiDAR-based semantic labeling. To cope with the low resolution and extreme asymmetry in the aspect ratio of the used LiDAR images, we propose a dedicated network block called *LiLaBlock*, which is inspired by the GoogLeNet inception modules of [31]. The block structure is illustrated in Fig. 3. In order to successfully handle relevant objects of various aspect ratios, the *LiLaBlock* applies convolution kernels of sizes $7 \times 3$, $3 \times 7$ and $3 \times 3$ in parallel. The output is then concatenated and the dimension is decreased by a factor of three via a bottleneck. In this way the dimensionality of the feature space is reduced yielding a more compact representation. At the same time the inference complexity of the *LiLaNet* is lowered. Note that each convolution is followed by a rectified linear unit layer (ReLU layer).

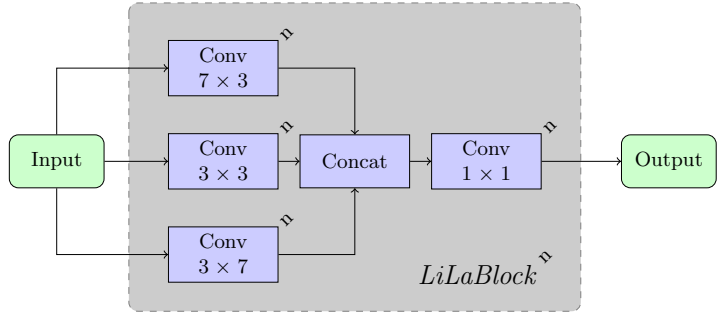

**Fig. 3.** The *LiLaBlock* structure allows to cope with the extreme asymmetry in the aspect ratio of the input LiDAR images.

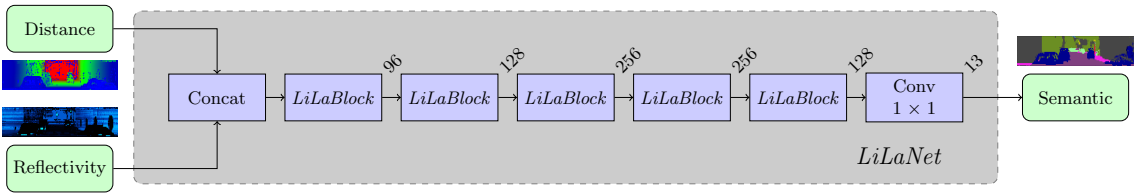

**Fig. 4.** The *LiLaNet* consists of a sequence of five consecutive *LiLaBlocks*. The final $1 \times 1$ convolution reduces the dimensionality to the desired label set.

The full *LiLaNet* consists of a sequence of five *LiLaBlocks* with a varying number of kernels, as shown in Fig. 4. The two input channels represent the concatenated depth and reflectivity images, while the output provides the corresponding point-wise semantic image labeling, according to the label set defined in Section 3.2.

The network training is performed via the Adam solver [32]. We use the suggested default values of $\beta_1 = 0.9$, $\beta_2 = 0.999$ and $\epsilon = 10^{-8}$. The learning rate is fixed at $\alpha = 10^{-3}$ ($\alpha = 10^{-4}$ for fine-tuning) and the batch size is set to $b = 5$, while the weights are initialized with MSRA [33].

### 3.4 Autolabeling

Generating manually annotated training data for LiDAR-based semantic labeling at scale presents a huge effort and entails even higher cost compared to manual image annotations in the 2D domain. This is due to both the additional spatial dimension and the sparsity of the data, which results in a representation that is non-intuitive and cumbersome for human annotators. For these reasons, we introduce an efficient automated process for large-scale training data generation called *Autolabeling*.

As illustrated in Fig. 5, the *Autolabeling* concept is based on the use of one or more reference cameras in conjunction with the LiDAR sensor capturing the point cloud data. The obtained reference camera images have to be registered to the LiDAR data in space and time. Preferably, the spatial displacement between the sensor origins is minimized to avoid occlusion artifacts.

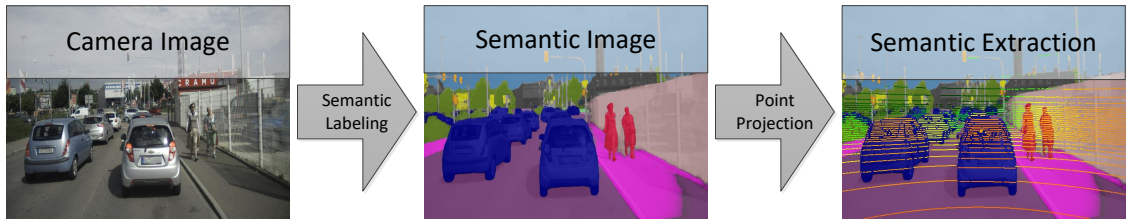

**Fig. 5.** Overview of the *Autolabeling* process for large-scale automated training data generation. Each pixel of the reference camera image is classified via an image-based semantic labeling network. Subsequently, the point cloud is projected into the camera image and the reference labels are transfered to the corresponding LiDAR points.

In the first step, a high-quality pixel-wise semantic labeling of the reference camera image is computed via state-of-the-art deep neural networks, as can be found on the leaderboard of the Cityscapes benchmark [1]. Second, the captured point cloud is projected into the reference image plane to transfer the semantic information of the image pixels to the corresponding LiDAR points. While a single reference camera will in general only cover a fraction of the full point cloud, it is straightforward to extend the approach to multiple cameras for increased coverage.

The described fully automated procedure yields semantically labeled point clouds which can directly be used to train LiDAR-based semantic labeling networks such as *LiLaNet*. In the following subsections we describe the various stages of the data generation process in more detail.

**Semantic Image Labeling** For the experiments in the present paper we use an efficient reference network as described in [34] to obtain the pixel-wise semantic labeling of the camera images. The network is trained on the Cityscapes dataset and achieves an Intersection-over-Union (IoU) test score of 72.6% with regard to the original Cityscapes label set. Since the Cityscapes dataset was designed with vehicle-mounted front-facing cameras in mind, we also use a single front-facing camera to evaluate the proposed automated training data generation process.

Note that the *Autolabeling* process can be applied using any image-based reference network providing sufficient output quality. Moreover, the process will in general directly benefit from the ongoing research and improvements in image-based semantic labeling networks.

**Point Projection** In order to project the 3D points captured by a scanning LiDAR into the reference camera image plane, several aspects have to be taken into account. Since the LiDAR scanner rotates around its own axis in order to obtain a 360° point cloud, each measurement is taken at a different point in time. In contrast, the camera image is taken at a single point in time, or at least with a comparatively fast shutter speed.

To minimize potential adverse effects introduced by the scanning motion of the LiDAR, we apply a point-wise ego-motion correction using vehicle odometry.

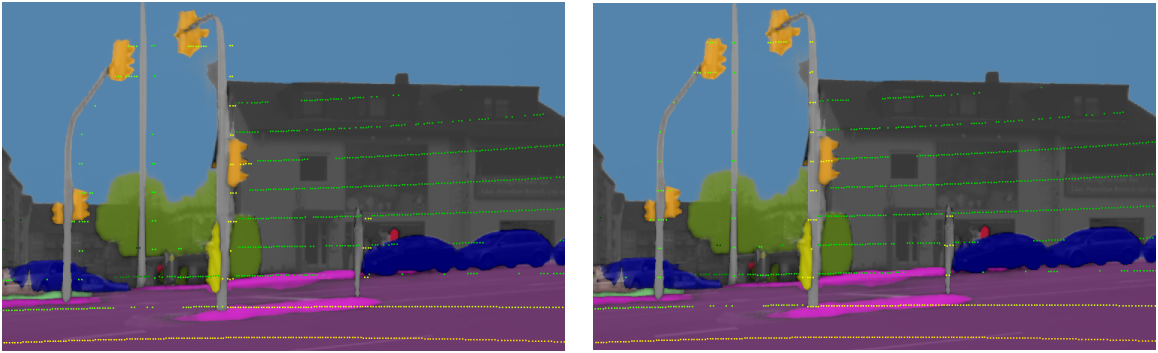

**Fig. 6.** Example of the point projection with (right) and without (left) ego-motion correction. The colors of the semantically labeled images represent the Cityscapes label set, while the projected points are color-coded according to distance (dark green = far away, yellow = close). Note that invalid LiDAR points are not projected into these images.

First, the measured 3D points are transformed from the LiDAR coordinate system to the vehicle coordinate system via the extrinsic calibration parameters of the sensor. Given the points $\boldsymbol{p_v} = (x_v, y_v, z_v)$ in the vehicle coordinate system, the wheel odometry data is used to compensate for the ego-motion of the vehicle. To this end, the time difference $\Delta t$ between the point measurement timestamp $t_p$ and the image acquisition timestamp $t_c$ of the camera is computed for each point. In case of a rolling shutter camera, half of the shutter interval $t_r$ is added to move the reference timestamp to the image center:

$$\Delta t = t_c - t_p + \frac{t_r}{2} \tag{1}$$

The time difference $\Delta t$ is used to extract the corresponding ego-motion data of the vehicle from the odometry sensor. This yields a transformation matrix $\boldsymbol{T_{\Delta t}}$ describing the motion that occurred between the two timestamps of interest. Using the transformation matrix $\boldsymbol{T_{\Delta t}}$, each point $\boldsymbol{p_v}$ is ego-motion corrected with

$$\begin{bmatrix} \boldsymbol{p_{t_c}} \\ 1 \end{bmatrix} = \begin{bmatrix} x_{t_c} \\ y_{t_c} \\ z_{t_c} \\ 1 \end{bmatrix} = \boldsymbol{T_{\Delta t}^{-1}} * \begin{bmatrix} x_v \\ y_v \\ z_v \\ 1 \end{bmatrix} \tag{2}$$

This effectively transforms each point $\boldsymbol{p_v}$ in the vehicle coordinate system to its corresponding position $\boldsymbol{p_{t_c}} = (x_{t_c}, y_{t_c}, z_{t_c})$ at camera timestamp $t_c$. Finally, the corrected points $\boldsymbol{p_{t_c}}$ are transformed to the camera coordinate system and projected to the image plane using a pinhole camera model [35]. An exemplary result of the point projection with and without ego-motion correction is shown in Fig. 6.

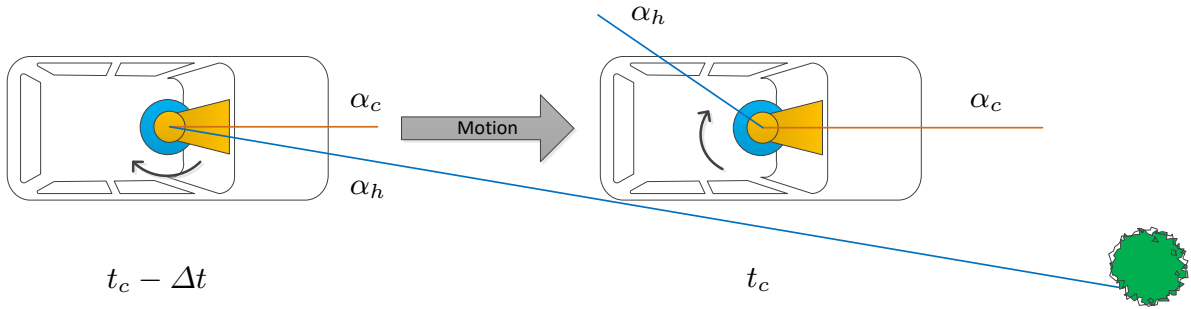

**Fig. 7.** Illustration of the timing problem (top view), where the camera (orange) is mounted on top of the LiDAR (blue). At a given timestamp, the rotating LiDAR measures points only at the angle $\alpha_h$ (blue line). The tree (green) is detected by the LiDAR at timestamp $t_c - \Delta t$, while the camera captures an image of the tree at $t_c$. This time difference $\Delta t$ results in a spatial displacement of the sensor origins, which causes occlusion artifacts resembling a large mounting distance between the two sensors.

|  | Training | Validation | Testing |
|---|---|---|---|
| Original Frames | 344,027 | 73,487 | 137,682 |
| Optimized Frames | 57,315 | 12,259 | 22,964 |
| Manual Annotated Keyframes | 1,909 | 373 | 718 |

**Table 2.** Split of the data into the different datasets for training, validation and testing.

## 4   Dataset

To train *LiLaNet*, a large dataset was created based on the *Autolabeling* process described Section 3. Approximately 555,000 frames of different road types (cities, rural roads, highways) were recorded and automatically labeled. Subsequently the recorded frames were split into subsets for training, validation and testing, with a split of approximately 62% - 13% - 25%. Details are listed in Table 2. The dataset split was performed on a sequence basis instead of via random selection in order to prevent correlations in between subsets. The training dataset was used to train *LiLaNet*, while the validation dataset was used to optimize the network architecture (not shown within this publication). The testing dataset is used to evalute the achieved LiDAR-based semantic labeling performance and the benefits of the proposed *Autolabeling* process.

An essential factor for accurate *Autolabeling* results is a small spatial distance between the LiDAR sensor and the reference camera, as occlusion artifacts tend to introduce inconsistencies to the datasets. This can be accommodated by both a small mounting distance of the two sensors and explicitly taking time synchronization into account. In the used setup, the camera is not being triggered by the LiDAR sensor, which results in cases where the LiDAR sensor orientation is not well aligned with the front-facing camera during image acquisition time. This misalignment causes a significant timestamp difference between the projected point cloud and the image data. While this effect is compensated by the ego-motion correction described in Section 3.4, it usually results in a significant translatory motion between the two sensors while driving, which in turn leads to considerable occlusion artifacts. See Fig. 7 for an illustration of the problem. For this reason, we define a maximum heading deviation range $\gamma_h = 60°$ which is allowed between the camera principal axis orientation $\alpha_c$ and the current LiDAR azimuth angle $\alpha_h$ at the camera time stamp $t_c$. All captured frames which lie outside of this maximum deviation range are discarded, yielding an optimized subset of the originally recorded data (see Table 2).

Based on the optimized datasets, 3,000 keyframes were selected equally distributed across all frames for manual annotation. This manually annotated data forms the basis for the evaluation in Section 5. Note that invalid LiDAR points are not being annotated and hence the class 'sky' is not considered for the evaluation in the present paper.

## 5   Experiments

The *Autolabeling* process introduced in Section 3.4 was applied to automatically generate the large-scale dataset described in Section 4, which in turn was used to train the proposed *LiLaNet* architecture for LiDAR based semantic labeling. In this section we evaluate different training strategies for *LiLaNet* in detail. In particular, we analyze the impact of boosting the training via the large-scale datasets obtained from the *Autolabeling* process. Note that all evaluations are based on the testing subset of the manually annotated keyframes. Following the Cityscapes Benchmark Suite [1], we apply the Intersection-over-Union (IoU) metric for performance evaluation. We define the following evaluation schemes, which are also visualized in Fig. 8:

(1) **Autolabeling:**
This evaluation considers inaccuracies of the image CNN as well as projection errors. No LiDAR-based CNN is trained for this evaluation.
(2) ***LiLaNet* Manual Annotations:**
This evaluation assesses the performance of the LiDAR-based semantic labeling using a small set of cost-intensive manually annotated point clouds from the annotated keyframes for training (see Table 2).
(3) ***LiLaNet* Autolabeled Reduced:**
This evaluation measures the performance of the LiDAR-based semantic labeling using a small set of automatically generated training data, based on the annotated keyframes (same keyframes that constitute the training subset of the manually annotated dataset shown in Table 2).
(4) ***LiLaNet* Autolabeled Full:**
This evaluation assesses the performance of the LiDAR-based semantic labeling when using the *Autolabeling* process on the full training dataset of the original frames (see Table 2).
(5) ***LiLaNet* Finetuned:**
This evaluation measures the performance of the LiDAR-based semantic labeling by fine-tuning the network using a small set of manually annotated data (2) with a pre-training based on the full *Autolabeling* process (4).

**Labeling Type**

|  | Manual Annotated | Autolabeled |
|---|---|---|
| **Manual Annotated Keyframes** | Manual Annotations (2) | Autolabeled Reduced (3) |
| **All Frames** | Does not exist (only a few frames are manual annotated) | Autolabeled Full (4) |

*Amount of Training Data*

**Fig. 8.** Overview of the various training strategies, which differ in the amount of training data (see Table 2) as well as the labeling type of the ground truth.

| | road | sidewalk | person | rider | small vehicle | large vehicle | two wheeler | construction | pole | traffic sign | vegetation | terrain | mean IoU |
|---|---|---|---|---|---|---|---|---|---|---|---|---|---|
| Image Labeling [34] based on Cityscapes | *98.0%* | *81.5%* | *81.4%* | *61.3%* | *94.7%* | *80.8%* | *70.2%* | *91.8%* | *58.0%* | *71.8%* | *92.7%* | *69.0%* | *79.3%* |
| (1) *Autolabeling* no CNN | 89.2% | 60.2% | **75.3%** | **51.3%** | 79.3% | **57.3%** | 45.0% | 68.5% | 28.9% | 39.4% | 78.0% | 53.2% | 60.5% |
| (2) *LiLaNet* Manual Annotations | 90.8% | 61.6% | 48.8% | 15.2% | 79.7% | 37.4% | 22.4% | 71.1% | 35.9% | 69.4% | 75.1% | 59.9% | 55.6% |
| (3) *LiLaNet* Autolabeled Reduced | 86.8% | 51.3% | 44.9% | 13.2% | 72.6% | 32.7% | 19.0% | 60.2% | 20.9% | 45.7% | 66.4% | 44.2% | 46.5% |
| (4) *LiLaNet* Autolabeled Full | 89.7% | 61.7% | 72.2% | 46.6% | 79.6% | 49.6% | 38.3% | 75.0% | 31.5% | 50.2% | 78.0% | 49.8% | 60.2% |
| (5) *LiLaNet* Finetuned | **94.1%** | **73.9%** | 73.8% | 48.9% | **86.4%** | 52.2% | **49.2%** | **83.4%** | **46.6%** | **75.7%** | **84.8%** | **67.4%** | **69.7%** |
| (4) SqueezeSeg [17] Autolabeled Full | 89.0% | 60.9% | 56.7% | 6.1% | 76.4% | 39.2% | 25.9% | 66.6% | 18.6% | 46.8% | 73.0% | 57.3% | 51.4% |
| (5) SqueezeSeg [17] Finetuned | 92.2% | 68.2% | 56.8% | 12.9% | 80.1% | 38.5% | 33.1% | 72.0% | 26.1% | 67.1% | 75.7% | 63.0% | 57.1% |

**Table 3.** Class-wise and overall IoU scores of the different approaches. The highest IoU scores of each column (excluding the first row) are marked in bold.

The used image-based semantic labeling network (Section 3.4) was previously analyzed in detail in [34], achieving an IoU score of 72.6% on the Cityscapes Benchmark Suite. In addition, we re-evaluate the image-based approach on the Cityscapes test set when using the class mapping defined in Section 3.2. This yields an IoU score of 79.3%, detailed results are listed in Table 3. Compared to the pure image-based semantic labeling result, the *Autolabeling* (1) output used to generate our large-scale dataset shows worse performance scores. This may be attributed to various reasons. First of all, the semantic image labeling CNN is optimized towards the specific properties of high-resolution camera data, which may lead to a slight reduction in performance when transferring the labels into the LiDAR domain. Also, despite the ego-motion correction efforts described in Section 4, the remaining inaccuracies still result in a certain amount of occlusion artifacts. Finally, a multi-sensor setup is rarely free from small calibration offsets in practice, even if optimized manually. These imperfections cause projection misalignments of the points within the images, which results in inaccurate label assignments, in particular for points at a large distance to the sensors.

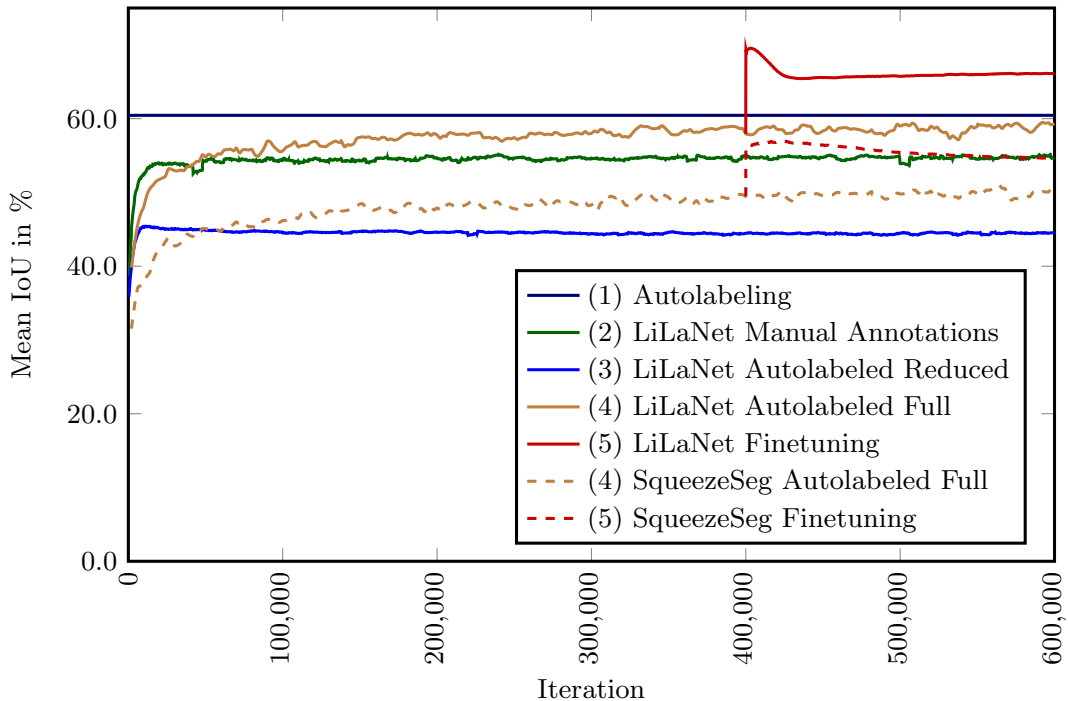

**Fig. 9.** Mean IoU of the different approaches and network architectures during training.

The detailed results of the various training strategies applied to *LiLaNet* are listed in Table 3 and illustrated in Fig. 9. It can be seen that the training on manually annotated data (2) yields a higher performance than the training on autolabeled data (3), but only as long as the same amount of data is being used. This is due to the imperfect results of the *Autolabeling* output itself. However, as stated previously, in practice the amount of available manually annotated data is severely limited by the high cost of point-wise manual annotation. In contrast, the *Autolabeling* process allows to automatically generate training data sets of arbitrary size at low cost. When using the large amount of automatically generated data for training (4), *LiLaNet* in fact outperforms its respective variant trained on manual annotations (2) by 4.6 percentage points with regard to mean IoU. Moreover, the network seems to generalize well within the LiDAR domain and suppresses some errors introduced by the *Autolabeling* process, which is indicated by the improved performance for some classes when compared to the pure *Autolabeling* output (e.g. 'construction', 'pole', 'traffic sign'). Furthermore, Fig. 9 shows that the training on the small amount of data in (2) and (3) saturates after several thousand iterations, while the training on the large-scale dataset (4) continues to increase output performance. Finally, using the manual annotations to fine-tune *LiLaNet* after pre-training on the automatically generated dataset (5) boosts performance by another 9.5 percentage points. This corresponds to a total gain of 14.1 percentage points over the training on manually annotated data only. Note that after fine-tuning most classes achieve a significantly higher performance than obtained by the pure *Autolabeling* output itself. Hence, the LiDAR-based semantic labeling result provided by *LiLaNet* with a training supported by the *Autolabeling* process outperforms the image-based semantic labeling results projected into the LiDAR domain. It is worth noting that the network fine-tuning reaches its maximum performance only after a few thousand iterations and soon starts to overfit on the small manually annotated dataset, as can be seen in Fig. 9. A qualitative result of the fine-tuned network is shown in Fig. 1.

In order to compare the presented *LiLaNet* architecture to the state-of-the-art, we also analyze the performance of the SqueezeSeg architecture [17], recently proposed for LiDAR-based semantic labeling on a smaller set of semantic classes ('car', 'pedestrian', 'cyclist'). Note that we evaluate the SqueezeSeg approach without its Conditional Random Field (CRF) stage for a fair comparison. The results in Table 3 as well as Fig. 9 illustrate that *LiLaNet* outperforms the SqueezeNet architecture in each class. This may be due to the following reasons: First of all, SqueezeNet uses five horizontal pooling layers which increases the learnable feature size drastically. Consequently, the network may have difficulties in capturing small but relevant objects. Further, the SqueezeSeg architecture does not distinguish between different object shapes and sizes in the design of the convolution kernels, as is done in our *LiLaBlock* structure. However, note that SqueezeNet does indeed also benefit from our combined process of fine-tuning after pre-training on an automatically generated dataset.

## 6    Conclusion

Autonomous vehicles require a comprehensive and robust environment representation, including a detailed semantic understanding of their surroundings. This can be obtained by a combination of different sensor modalities, where each sensor independently contributes to the overall environmental model. In the present paper we consider the point-wise multi-class semantic labeling of 3D point clouds and transfer the concept of pixel-wise image-based semantic labeling to the LiDAR domain. We propose *LiLaNet*, a novel CNN architecture for efficient LiDAR-based semantic labeling. This architecture significantly outperforms current state-of-the-art CNNs for multi-class LiDAR-based semantic labeling when evaluated on a manually annotated ground truth dataset.

Furthermore, we present a fully automated process for large-scale cross-modal training data generation called *Autolabeling*. The approach is based on the use of reference cameras in order to transfer high-quality image-based semantic labeling results to LiDAR point clouds. Combining the automatically generated training dataset with a fine-tuning step based on small-scale manually annotated data yields a performance boost of up to 14 percentage points while keeping manual annotation efforts low.